\documentclass{article}
\usepackage{spconf,amsmath,graphicx}
\usepackage{algorithm,algpseudocode}
\usepackage{mathtools}
\usepackage{cancel}
\usepackage{amssymb}
\usepackage{booktabs}
\usepackage{caption}
\usepackage{subcaption}
\usepackage{multirow}
\usepackage{setspace}
\usepackage{icomma}

\usepackage{pifont}
\usepackage{tikz}
\usetikzlibrary{positioning}
\usetikzlibrary{calc}
\usepackage{pgfplots, pgfplotstable}    
\pgfplotsset{compat=1.17} 
\usepackage{url}
\DeclareMathSymbol{\shortminus}{\mathbin}{AMSa}{"39}

\makeatletter
\def\blfootnote{\xdef\@thefnmark{}\@footnotetext}
\makeatother

\usepackage[
backend=biber,
style=ieee,
citestyle=numeric-comp,
maxbibnames=3,
maxcitenames=3,
doi=false,isbn=false,url=false,eprint=false
]{biblatex}

\defbibheading{bibliography}[\refname]{}

\usepackage[
backend=biber,
style=ieee,
citestyle=numeric-comp,
maxbibnames=3,
maxcitenames=3,
doi=false,isbn=false,url=false,eprint=false
]{biblatex}

\DeclareSourcemap{
	\maps[datatype=bibtex, overwrite=true]{
		\map{
			\step[fieldsource=booktitle,
			match=\regexp{.*Interspeech.*},
			replace={Proc. Interspeech}]
			\step[fieldsource=journal,
			match=\regexp{.*INTERSPEECH.*},
			replace={Proc. Interspeech}]
			\step[fieldsource=booktitle,
			match=\regexp{.*ICASSP.*},
			replace={Proc. ICASSP}]
			\step[fieldsource=booktitle,
			match=\regexp{.*icassp_inpress.*},
			replace={Proc. ICASSP (in press)}]
			\step[fieldsource=booktitle,
			match=\regexp{.*Acoustics,.*Speech.*and.*Signal.*Processing.*},
			replace={Proc. ICASSP}]
			\step[fieldsource=booktitle,
			match=\regexp{.*International.*Conference.*on.*Learning.*Representations.*},
			replace={Proc. ICLR}]
			\step[fieldsource=booktitle,
			match=\regexp{.*International.*Conference.*on.*Computational.*Linguistics.*},
			replace={Proc. COLING}]
			\step[fieldsource=booktitle,
			match=\regexp{.*SIGdial.*Meeting.*on.*Discourse.*and.*Dialogue.*},
			replace={Proc. SIGDIAL}]
			\step[fieldsource=booktitle,
			match=\regexp{.*International.*Conference.*on.*Machine.*Learning.*},
			replace={Proc. ICML}]
			\step[fieldsource=booktitle,
			match=\regexp{.*North.*American.*Chapter.*of.*the.*Association.*for.*Computational.*Linguistics:.*Human.*Language.*Technologies.*},
			replace={Proc. NAACL}]
			\step[fieldsource=booktitle,
            match=\regexp{.*EMNLP.*},
			replace={Proc. EMNLP}]
			\step[fieldsource=booktitle,
			match=\regexp{.*Empirical.*Methods.*in.*Natural.*Language.*Processing.*},
			replace={Proc. EMNLP}]
			\step[fieldsource=booktitle,
			match=\regexp{.*Association.*for.*Computational.*Linguistics.*},
			replace={Proc. ACL}]
			\step[fieldsource=booktitle,
			match=\regexp{.*Automatic.*Speech.*Recognition.*and.*Understanding.*},
			replace={Proc. ASRU}]
			\step[fieldsource=booktitle,
			match=\regexp{.*Spoken.*Language.*Technology.*},
			replace={Proc. SLT}]
			\step[fieldsource=booktitle,
			match=\regexp{.*Speech.*Synthesis.*Workshop.*},
			replace={Proc. SSW}]
			\step[fieldsource=booktitle,
			match=\regexp{.*workshop.*on.*speech.*synthesis.*},
			replace={Proc. SSW}]
			\step[fieldsource=booktitle,
			match=\regexp{.*Advances.*in.*neural.*information.*processing.*},
			replace={Proc. NeurIPS}]
			\step[fieldsource=booktitle,
			match=\regexp{.*Advances.*in.*Neural.*Information.*Processing.*},
			replace={Proc. NeurIPS}]
			\step[fieldsource=booktitle,
			match=\regexp{.*Workshop.*on.* Applications.* of.* Signal.*Processing.*to.*Audio.*and.*Acoustics.*},
			replace={Proc. WASPAA}]
			\step[fieldsource=publisher,
			match=\regexp{.+},
			replace={{}}]
			\step[fieldsource=month,
			match=\regexp{.+},
			replace={{}}]
			\step[fieldsource=location,
			match=\regexp{.+},
			replace={{}}]
			\step[fieldsource=address,
			match=\regexp{.+},
			replace={{}}]
			\step[fieldsource=organization,
			match=\regexp{.+},
			replace={{}}]
		}
	}
}

\title{A Study on the Integration of Pipeline and E2E SLU systems for Spoken Semantic Parsing toward STOP Quality Challenge}
\name{
\begin{tabular}{c}
Siddhant Arora${}^1$, Hayato Futami${}^2$, Shih-Lun Wu${}^1$,Jessica Huynh${}^1$,\\ Yifan Peng${}^1$,Yosuke Kashiwagi${}^2$, Emiru Tsunoo${}^2$, Brian Yan${}^1$, Shinji Watanabe${}^1$
\end{tabular}
}
\address{${}^1$Carnegie Mellon University, ${}^2$Sony Group Corporation, Japan}
\begin{document}
\maketitle
\begin{abstract}
Recently there have been efforts to introduce new benchmark tasks for spoken language understanding (SLU), like semantic parsing. In this paper, we describe our proposed spoken semantic parsing system for the quality track (Track 1) in Spoken Language Understanding Grand Challenge which is part of ICASSP Signal Processing Grand Challenge 2023. We experiment with both end-to-end and pipeline systems for this task. Strong automatic speech recognition (ASR) models like Whisper and pretrained Language models (LM) like BART are utilized inside our SLU framework to boost performance. We also investigate the output level combination of various models to get an exact match accuracy of 80.8, which won the 1st place at the challenge.
\end{abstract}
\section{Introduction}
\label{sec:intro}
Spoken Language Understanding Grand Challenge or Spoken Task Oriented Parsing (STOP) Challenge, which is part of ICASSP Signal Processing Grand Challenge 2023, aims to build systems that can convert a spoken utterance to a semantic parse sequence to facilitate the execution of tasks by the voice assistant. This work discusses our team PittOsaki's approach for Track 1 to improve the quality of generated semantic parse using open source model and ASR datasets. 

In this work, we experiment with various pipeline, and end-to-end (E2E) SLU approaches. Pretrained self-supervised speech (SSL) representations like WavLM and Hubert are employed in our SLU framework. We also incorporate pretrained LMs like BART large. Finally, a system combination of various models shows a significant performance gain over the baseline systems.

\section{METHODOLOGY}
We formulate the SLU task of semantic parsing as a unified sequence-to-sequence problem. The input is a sequence of speech features extracted from the raw audio, and the output is a semantic parse represented as a linearized tree structure. The attention-based encoder-decoder architecture is adopted in our end-to-end (E2E) SLU approaches. Our ASR model is also based on encoder-decoder architecture. For both ASR and SLU training, we employ SSL representations like WavLM as a frontend. A weighted sum of multiple hidden states is utilized and the parameters are frozen during training. We also experiment with utilizing the recently released Whisper model in our SLU framework. The entire Whisper model is fine-tuned instead of it being used as a frontend since it achieved superior performance in our initial experimentations. 
Our NLU model is incorporated by fine-tuning pre-trained LMs.
Similar to prior work~\cite{Ddel1}, we improve the semantic modeling of our E2E SLU models by adopting a 2-pass SLU approach~\cite{2-pass-slu}, where the second pass combines both acoustic and semantic information generated by pretrained LM from ASR hypotheses. Inspired by the principles of task compositionality, we also train compositional E2E SLU model~\cite{arora-etal-2022-token} that first convert the spoken utterance to a sequence of token representations, which can then be used in the traditional NLU framework. 
To combine hypotheses from multiple models, we directly apply the recognizer output voting error reduction (ROVER) method and extract exact match (EM) accuracy from the combined semantic parse sequence.

\section{Experiment Setup}
\label{sec:exp_setup}
We adopt the evaluation metrics in the STOP benchmark i.e. EM accuracy. The encoder of our E2E SLU model is a 12-layer Conformer, while the decoder is a 6-layer Transformer. The number of heads and dimension of a self-attention layer is set to 8 and 512. The linear units are 2048 for the encoder and the decoder. Speed perturbation and SpecAugment are performed for data augmentation. Our ASR model has the same architecture as our E2E SLU model. We investigate using external ASR datasets like Librispeech and Commonvoice for pretraining. For 2 pass SLU models, our deliberation encoder consists of 4-layer conformer, and second-pass decoder has the same architecture as the ASR decoder. Our Compositional model consists of the same architecture as the ASR model in it’s ASR component and 6-layer transformer encoder, and 6-layer transformer decoder in it’s NLU component. Dropout and label smoothing are applied. 
For pretrained LMs, we use BART large and T5 large, which are trained with HuggingFace Seq2Seq Trainer. We add special tokens in the vocabulary for slot and intent tags. 

We also experiment with Whisper medium for ASR and both Whisper medium and large models for SLU. For ASR, we train in 3 settings: (a) using lowercase transcripts (b) transcripts with original casing (as in ``utterance'' field) referred to as Whisper w/ casing (c) original casing transcripts and more frequent saving of checkpoints (i.e. after 1k iterations instead of epoch) referred to as Whisper w/ freq. checkpoint. Similar to pretrained LMs, we add special tokens in vocabulary for slot and intent tags while training E2E SLU models. 
More details about our models and the config files will be publicly available as part of the ESPnet-SLU~\cite{ESPnet-SLU} toolkit.
\begin{table}[t]
\centering
\resizebox {0.9\linewidth} {!} {
\begin{tabular}{ll|c}
\toprule
& Pre-trained Model & Test EM ($\uparrow$) \\ \midrule
\multicolumn{3}{l}{\textbf{STOP benchmark}~\cite{stop2022}} \\
\quad E2E & Wav2Vec2 & 68.7\hphantom{0}\\  
\quad E2E & Hubert & 69.23 \\\midrule %
\quad Pipeline & Wav2Vec2+BART-L & 72.36 \\ %
\midrule
\midrule
\multicolumn{3}{l}{\textbf{Our E2E approach}} \\
\quad w/ SSL 
& Hubert  & 71.6\hphantom{0} \\
& WavLM  & 73.3\hphantom{0} \\\midrule
\quad w/ 2 pass SLU \\
& WavLM+BART-L & 75.8\hphantom{0} \\\midrule
\quad w/ Compositional E2E SLU \\
& WavLM  & 75.9\hphantom{0} \\\midrule
\midrule
\multicolumn{3}{l}{\textbf{Our Pipeline approach}} \\
& WavLM+BART-L  & 78.2\hphantom{0} \\
\bottomrule
\end{tabular}
}
\caption{Exact Match (EM) accuracy for semantic parsing on STOP dataset.}
\label{tbl:main-result}
\vskip -0.1in
\end{table}
\begin{table}[t]
\centering
\resizebox {0.9\linewidth} {!} {
\begin{tabular}{ll|c}
\toprule
& Pre-trained Model & Test EM ($\uparrow$) \\ \midrule
\midrule
\multicolumn{3}{l}{\textbf{Our E2E approach}} \\
\quad w/ SSL 
& Whisper  & 78.8\hphantom{0} \\
& Whisper large  & 79.1\hphantom{0} \\\midrule
\quad w/ 2 pass SLU  \\
& w/ Whisper transcripts & 77.0\hphantom{0} \\\midrule
\quad w/ Compositional E2E SLU \\
& w/ Whisper transcripts  & 77.4\hphantom{0} \\\midrule
\midrule
\multicolumn{3}{l}{\textbf{Our Pipeline approach}} \\
& Best ASR+BART-L  & 80.5\hphantom{0} \\
& Best ASR+T5  & 80.1\hphantom{0} \\\midrule
\midrule
\multicolumn{3}{l}{\textbf{Our System Combination}} \\
& 4 best models  & 80.8\hphantom{0} \\
\bottomrule
\end{tabular}
}
\caption{Exact Match (EM) accuracy for semantic parsing using Whisper on STOP dataset.}
\label{tbl:main-whisper-result}
\vskip -0.2in
\end{table}
\begin{table}[t]
\centering
\resizebox {\linewidth} {!} {
\begin{tabular}{ll|c}
\toprule
Pre-trained Model & Pretrained Dataset & Test WER ($\downarrow$) \\ \midrule
\multicolumn{3}{l}{\textbf{STOP benchmark}~\cite{stop2022}} \\
\quad Wav2Vec2 & STOP & 4.45\\  
\quad Hubert & STOP & 4.26 \\\midrule %
\midrule
\multicolumn{3}{l}{\textbf{Our ASR models}} \\
\quad Hubert  & STOP & 3.8\hphantom{0} \\
\quad WavLM  & STOP & 3.3\hphantom{0} \\
\quad WavLM w/ LM & STOP & 3.1\hphantom{0} \\
\quad WavLM w/ LM & Librispeech+Commonvoice+STOP & 2.7\hphantom{0} \\
\quad Whisper  & STOP & 2.4\hphantom{0} \\
\quad Whisper w/ casing  & STOP & 2.3\hphantom{0} \\
\quad Whisper w/ freq. checkpoint  & STOP & 2.3\hphantom{0} \\\midrule
\midrule
\multicolumn{3}{l}{\textbf{Our System Combination}} \\
& 4 best models  & 2.2\hphantom{0} \\
\bottomrule
\end{tabular}
}
\caption{Word Error Rate (WER) on STOP dataset.}
\label{tbl:asr-result}
\vskip -0.2in
\end{table}
\section{Results}
\label{sec:main_res}
Our results on the semantic parsing task without using the Whisper model are shown in Table~\ref{tbl:main-result}. 2-pass and compositional E2E SLU models perform better than traditional E2E models. We further observe that the pipeline model is significantly better than all the E2E SLU models. 
Table~\ref{tbl:main-whisper-result} shows that finetuning Whisper 
can be very helpful in improving E2E SLU performance since Whisper has been trained on large amounts of labeled data. Table~\ref{tbl:asr-result} similarly shows that Whisper achieves very good improvement on the ASR task. We further experiment with using Whisper transcripts in our 2 pass and compositional E2E SLU model directly during inference and observe significant performance gains. We were not able to investigate incorporating the Whisper model in our Compositional and 2 pass E2E SLU model due to time constraints, but we will investigate this in future work.

Finally, we use the ROVER combination on the hypothesis produced by our 4 best ASR models (Table~\ref{tbl:asr-result}) and achieve WER 2.2. Using this ASR transcript, we are able to significantly boost the performance of our pipeline systems. We further use ROVER to combine the 4 best SLU models (Table~\ref{tbl:main-whisper-result}) to achieve EM 80.8.

\section{Acknowledgement}
This work used Bridges2 system at PSC and Delta system at NCSA through allocation CIS210014 from the Advanced Cyberinfrastructure Coordination Ecosystem: Services \& Support (ACCESS) program, which is supported by National Science Foundation grants \#2138259, \#2138286, \#2138307, \#2137603, and \#2138296. Jessica Huynh was supported by NSF Graduate Research Fellowship grants DGE1745016 and DGE2140739. The opinions expressed in this paper do not necessarily reflect those of that funding agency.
\section{References}
{
\printbibliography
}
\end{document}